\documentclass{llncs}
\usepackage{graphicx}
\usepackage{array,color}
\usepackage{multirow}
\usepackage{arydshln}
\usepackage{tabularx}

\begin{document}

\title{Deep Learning Super-Resolution Enables Rapid Simultaneous Morphological and Quantitative Magnetic Resonance Imaging}
\titlerunning{Super-Resolution MRI}  
%

\author{Akshay Chaudhari\inst{1}, Zhongnan Fang\inst{2},  Jin Hyung Lee\inst{3}, Garry Gold\inst{1}, and Brian Hargreaves\inst{1} }

\institute{Department of Radiology, Stanford University, Stanford CA \\
\email{\{akshaysc, gold, bah\}@stanford.edu}
\and
LVIS Corporation, Palo Alto, CA \\
\email{\{zhongnanf\}@gmail.com}
\and
Department of Neurology, Stanford University, Stanford CA \\
\email{\{ljinhy\}@stanford.edu}
}
%
%

\maketitle              

\begin{abstract}
Obtaining magnetic resonance images (MRI) with high resolution and generating quantitative image-based biomarkers for assessing tissue biochemistry is crucial in clinical and research applications. However, acquiring quantitative biomarkers requires high signal-to-noise ratio (SNR), which is at odds with high-resolution in MRI, especially in a single rapid sequence. In this paper, we demonstrate how super-resolution can be utilized to maintain adequate SNR for accurate quantification of the T$_2$ relaxation time biomarker, while simultaneously generating high-resolution images. We compare the efficacy of resolution enhancement using metrics such as peak SNR and structural similarity. We assess accuracy of cartilage T$_2$ relaxation times by comparing against a standard reference method. Our evaluation suggests that SR can successfully maintain high-resolution and generate accurate biomarkers for accelerating MRI scans and enhancing the value of clinical and research MRI.

\end{abstract}
%
\section{Introduction}

Magnetic resonance imaging (MRI) is an excellent non-invasive diagnostic tool to accurately assess pathologies in several anatomies. However, MRI is fundamentally constrained in optimizing for either high-resolution, high signal-to-noise ratio (SNR), or low scan durations. Enhancing one of the three outcomes necessarily degrades one or both of the others. Additionally, unlike other imaging modalities, MR images are qualitative in nature and do not directly correlate to the underlying tissue physiology. While quantitative MRI may help in assessing tissue biochemistry and longitudinal changes, biomarker accuracy is extremely sensitive to image SNR. Consequently, it is challenging to develop a single MRI method to produce high-resolution morphological images with high quantitative biomarker accuracy in a reasonable scan time, which is tolerable for patients and which ultimately limits cost of the procedure. 

\subsection{Background}

The double-echo in steady-state (DESS) pulse sequence can generate high-resolution images with diagnostic contrast as well as the quantitative biomarker of T$_2$ relaxation time, in only five-minutes of scan time \cite{chaudhari2017five}. The T$_2$ relaxation time has shown to be sensitive to collagen matrix organization and tissue hydration levels, and is useful for assessing degradation of tissues such as cartilage, menisci, tendons, and ligaments \cite{mosher2004cartilage}. DESS intrinsically produces two images with independent contrasts. The first echo of DESS (S$_1$) has a T$_1$/T$_2$ weighting while the second echo of DESS (S$_2$) has a high T$_2$ weighting. 

\begin{figure}[tbp]
\begin{center}
\includegraphics[width=4.8in]{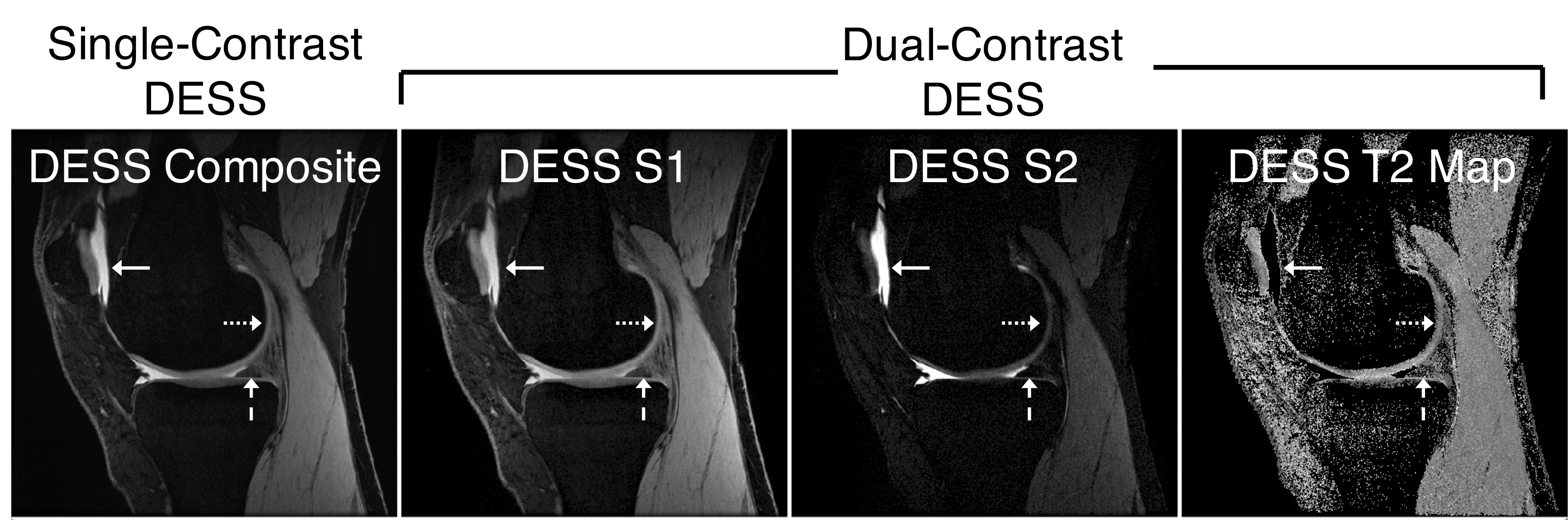} 
\end{center}
\vspace*{-5mm}
\caption{Compared to the single-contrast DESS, dual-contrast DESS provides additional morphological information and automatic quantitative T$_2$ relaxation time maps. The separate DESS contrasts (S$_1$ and S$_2$) and T$_2$ maps are useful in assessing the cartilage (dashed arrow), the menisci (dotted arrow), and inflammation (solid arrow). The T$_2$ maps are not affected by noisy fat-suppression of bony signal.}
\label{fig:dess}
\end{figure}

In previous applications of DESS, the S$_1$ and S$_2$ scans are combined during the reconstruction process to produce an output with a singular contrast (herein referred as \textit{single-contrast DESS}) \cite{Peterfy2008}. However, separating the two echoes can provide considerable diagnostic utility since both echoes are sensitive to varying pathologies. Additionally, the two independent-contrast images (herein referred as \textit{dual-contrast DESS}) can be used to analytically determine the tissue T$_2$ relaxation time, which is a promising biomarker for tissue degradation and OA progression \cite{mosher2004cartilage,Sveinsson2017}. Example images comparing the output of single-contrast DESS and dual-contrast DESS are shown in Fig.\ref{fig:dess}. Dual-contrast DESS has shown to be useful in diagnostic musculoskeletal imaging of knee as well as in research studies for evaluating OA progression \cite{chaudhari2017five,Monu2016}. 

\subsection{Motivation}

While promising, the dual-contrast DESS is limited in acquiring slices with 1.5mm section-thickness to maintain adequate SNR for T$_2$ measurements of the cartilage and menisci. Compared to an in-plane resolution of 0.4x0.4mm, such a high-section thickness precludes multi-planar reformations, which are essential for evaluating thin knee tissues in arbitrary planes, due to excessive image blurring. An ideal acquisition would provide sub-millimeter section thickness without biasing T$_2$ measurements. Advances in convolutional neural networks (CNNs) and 3D super-resolution (SR) methods may enable acquisition of slices with a thickness of 1.5mm followed by retrospectively achieving sub-millimeter resolution, while maintaining SNR for T$_2$ measurements \cite{vdsr}. However, unlike the single-contrast DESS that has hundreds of datasets publicly available, the dual-contrast DESS is a newer sequence with very limited amounts of high-resolution data available, which makes it challenging to create a SR CNN from scratch. In such scenarios, transfer learning methods may be helpful in overcoming the limitations of a paucity of high-resolution ground-truth dual-contrast DESS training data. Specifically, it may be possible to train a SR CNN initially using single-contrast DESS datasets and subsequently adapt the network to enhance dual-contrast DESS images using limited training data. 

Consequently, this study aimed to answer: \textbf{1.} Can transfer learning enhance through-plane MRI resolution for the clinically-relevant dual-contrast DESS sequence and \textbf{2.} Can transfer learning enable accurate quantitative imaging of the T$_2$ relaxation time by overcoming SNR limitations commonly faced in high-resolution imaging? The overall goal of this study was to evaluate whether there can be an efficient methodology to create a SR CNN for dual-contrast DESS to produce high-resolution morphological and quantitative images.
%
\section{Related Work}

Sparse-coding SR (ScSR) is a state-of-the-art non-deep-learning method that has been used for 2D MRI SR \cite{wang2014sparse}. CNN-based 3D SR MRI has previously shown to transform MRI images with a high section-thickness (low slice-direction resolution) into images with lower section-thickness (high slice-direction resolution) \cite{srcnn_blinded}. However, this initial training was performed on single-contrast DESS sequence that does not produce quantitative biomarkers. These scans were originally acquired with a section thickness of 0.7mm and retrospectively downsampled by a factor of 2x to a section thickness of 1.4mm to exactly duplicate a faster, lower-resolution acquisition. The SR network was then utilized to evaluate whether the original 0.7mm scans could be recovered from the 1.4mm slices. We build upon these results and to extend SR to MRI sequences that can simultaneously produce multiple diagnostic contrasts and quantitative biomarkers.

\section{Methods}

\subsection{Imaging Methodology}
We utilized a CNN termed Magnetic Resonance Super-Resolution (MRSR) to extend the SR capabilities of the network initially trained for single-contrast DESS scans. The dual-contrast DESS datasets used in this study were acquired with a slice thickness of 0.7mm (imaging parameters: TE$_1$/TE$_2$/TR = 7/39/23 ms, matrix size = 416x416, field of view = 160mm, flip angle = 20$^\circ$, scan time = 5 minutes, phase encoding parallel imaging = 2x, slices = 160). A slice thicknesses of 0.7mm was maintained for the single-contrast and dual-contrast DESS scans. 

A pre-trained network for performing SR with a slice downsampling factor of 2x for the single-contrast DESS sequence was utilized to simultaneously enhance both images from the dual-contrast DESS. This pre-training was performed on image patches with input and output sizes of 32x32x32 using convolutional filters of size 3x3x3 and a feature map length of 64. This SR CNN network transforms an input low-resolution image into a residual image through a series of 20 convolutions and rectified linear unit (ReLU) activations \cite{srcnn_blinded}. An approximate high-resolution image is generated through the sum of the low-resolution input and the resultant residual using the L2-norm between the approximate and true high-resolution images as the loss function. 

\subsection{Transfer Learning Training for Dual-Contrast DESS} 
\begin{figure}[tbp]
\begin{center}
\includegraphics[width=4.85in]{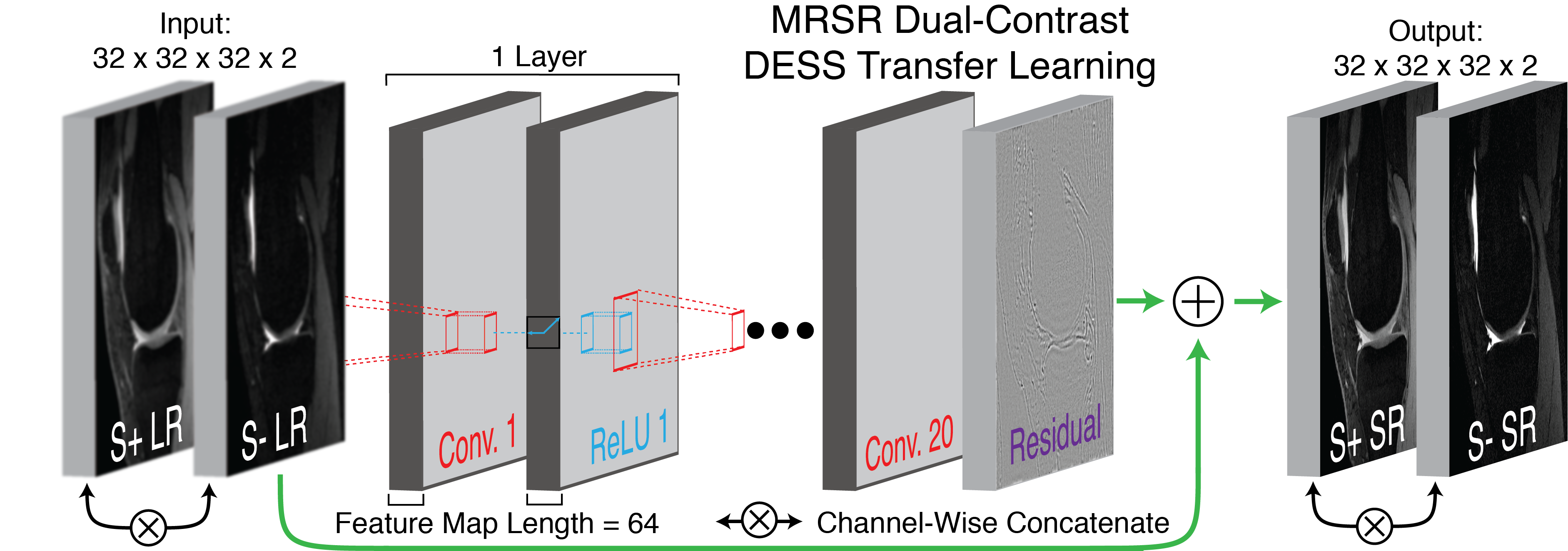} 
\end{center}
\vspace*{-5mm}
\caption[Schematic for the MRSR network]{The schematic of the Magnetic Resonance Super-Resolution (MRSR) network demonstrates how the low-resolution (LR) dual-contrast DESS images are simultaneously transformed into the super-resolution (SR) images.}
\label{fig:schematic}
\end{figure}

\vspace*{-2mm}
Since dual-contrast DESS contains an extra image contrast, the initial single-contrast DESS weights for the first convolution layer were duplicated to account for the dual-echoes. Similarly, the final layer output weights were modified to output two echo images instead of one, as shown in Fig. \ref{fig:schematic}. In such a manner, the single-contrast DESS MRSR architecture was modified and subsequently fine-tuned to simultaneously enhance dual-contrast DESS images.

All data processing steps for the single-contrast DESS and MRSR networks were were kept unchanged. This included data normalization between 0 and 1, simulation of thicker slices with a 48$^{th}$-order anti-aliasing filter, a mini-batch size of 50, and a learning rate of 0.0001. All input patches had a size of 32x32x32x2 with a stride of 16 in the first three directions. Thus, an input image of dimensions 416x416x160 was divided into 5625 patches. The MRSR network was trained for 10 epochs using 4 NVIDIA Titan 1080Ti graphical processing units.  

30 dual-contrast DESS 3D datasets were used for training and 10 for validation. All datasets were collected from patients referred for a clinical MRI following institutional review board approval and informed consent, for ensuring unbiased representation of healthy and pathologic tissues.

Two unique datasets, described below, were tested using the MRSR transfer learning network because it is not currently possible to acquire a single high-resolution dataset that also has high-SNR for accurate quantitative imaging of the T$_2$ relaxation time. The goal of this two-fold testing was to acquire separate reference high-resolution and high-SNR scans. The dual-contrast DESS could therefore have intermediate SNR for accurate T$_2$ measurements and the intermediate resolution of the acquisition could be enhanced using MRSR.

\vspace*{-6mm}
\subsubsection{Image Quality: Test Cohort 1}

This dataset had identical scan parameters to the training dataset. Following the simulation of 2x thicker slices, image quality enhancements were evaluated by comparing the structural similarity (SSIM), peak SNR (pSNR), and root mean square error (RMSE) between the ground truth high-resolution and MRSR  images, along with tricubic interpolated (TCI), Fourier interpolated (FI), and sparse coding super-resolution (ScSR) images.

\begin{figure}[tbp]
\begin{center}
\includegraphics[width=4.85in]{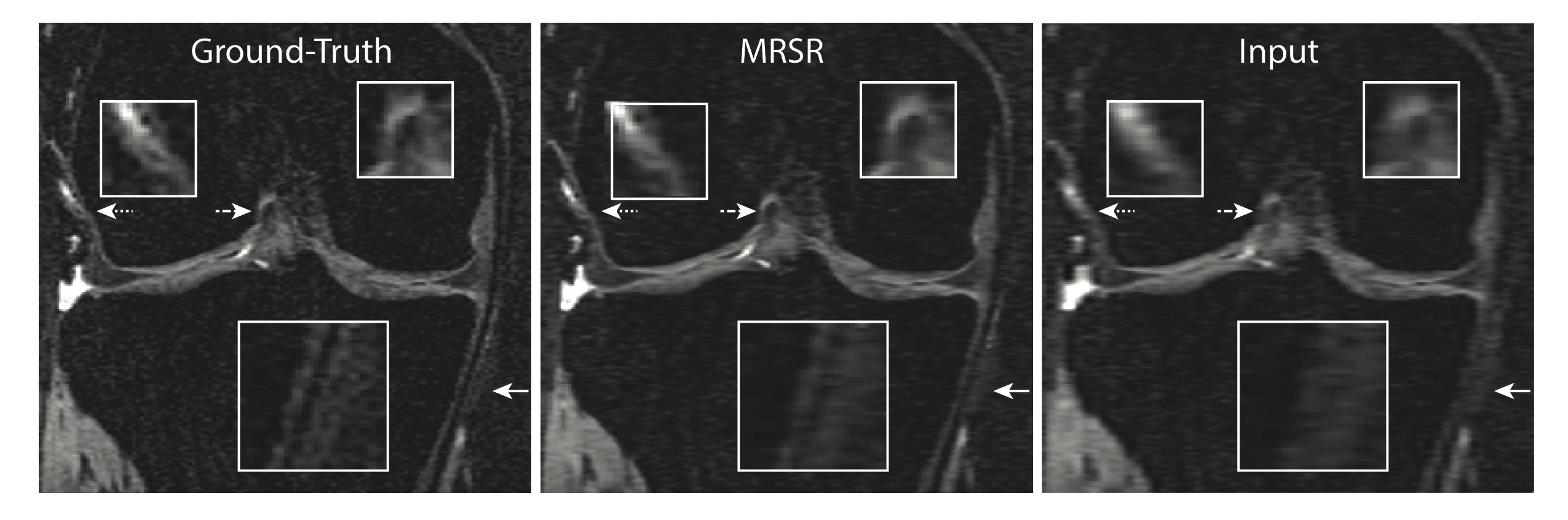} 
\end{center}
\vspace*{-5mm}
\caption[Example coronal MRSR echo S$_1$ images compared to the ground-truth and the MRSR input]{MRSR coronal reformatted images demonstrate better resolution in the slice-direction (left-right) than the input TCI images, compared to the ground-truth.}

\label{fig:cor}
\end{figure}
\vspace*{-0mm}

\begin{figure}[tbp]
\begin{center}
\includegraphics[width=4.85in]{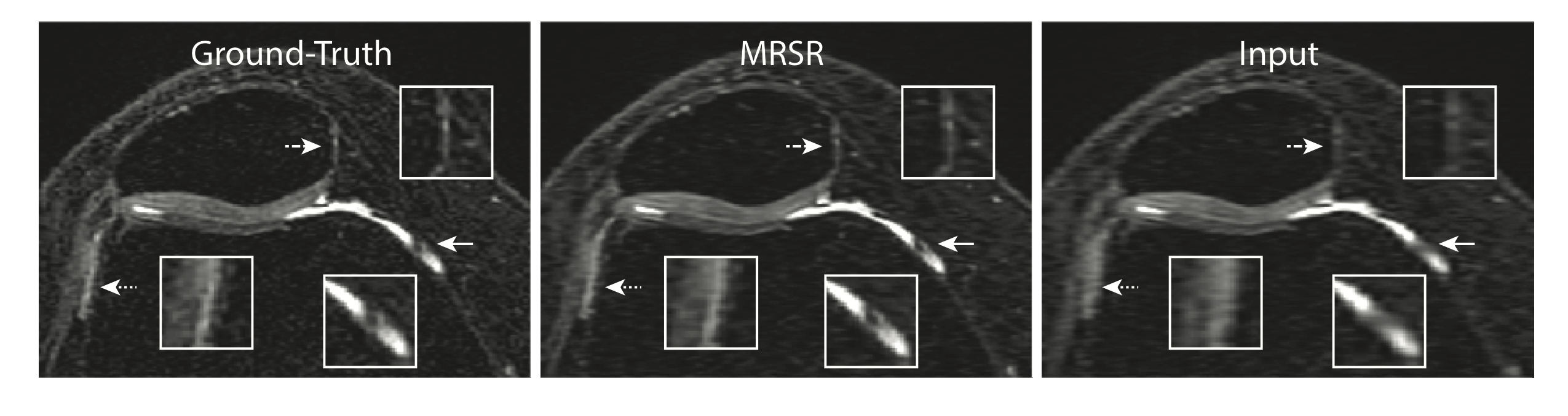} 
\end{center}
\vspace*{-5mm}
\caption[Example axial MRSR images compared to the ground-truth and the MRSR input]{Example axial reformatted MRSR images, depict finer image details considerably better than the input TCI image compared to the ground-truth.}
\label{fig:ax}
\end{figure}

\vspace*{-6mm}
\subsubsection{T$_2$ Accuracy: Test Cohort 2}

The second dataset had thicker slices (1.6mm) to maintain a higher SNR for accurate T$_2$ quantification, since T$_2$ has a high sensitivity to noise \cite{chaudhari2017five}. Accuracy of the T$_2$ maps was evaluated by comparing the T$_2$ values in two combined adjacent slices in the medial femoral cartilage of the MRSR, TCI, FI, and ScSR outputs to the ground-truth thick-slice sequences. Segmentation was performed by a reader with 5 years of experience in knee MRI segmentation. T$_2$ relaxation time differences, coefficients of variation (CV\%), and concordance correlation coefficients (CCC) assessed  T$_2$ variations between the methods, compared to the ground truth. 

Mann-Whitney U-Tests assessed variations between morphological enhancement metrics as well as T$_2$ variations for all enhancement methods. 

\section{Results}

Each epoch training duration was approximately 3 hours for the total of 170,000 training patches. The SSIM, pSNR, and RMSE values between the MRSR, TCI, FI, and ScSR images to the ground-truth are shown in Table \ref{tab:morph}, where MRSR was significantly superior compared to TCI, FI, and ScSR. Comparisons for T$_2$ values computed with all methods are shown in Table \ref{tab:t2}. MRSR had the best image quality metrics, as well as the closest matches for the T$_2$ values. Despite being compared on a pixel-wise basis, which can have a high sensitivity to noise, the MRSR T$_2$ values had the lowest inter-method CV of 3\% and an excellent CCC of 0.93. There were no statistically significant variations for T$_2$ for any method compared to the ground truth, likely due to a limited sample size. 

\begin{figure}[bp]
\begin{center}
\includegraphics[width=4.85in]{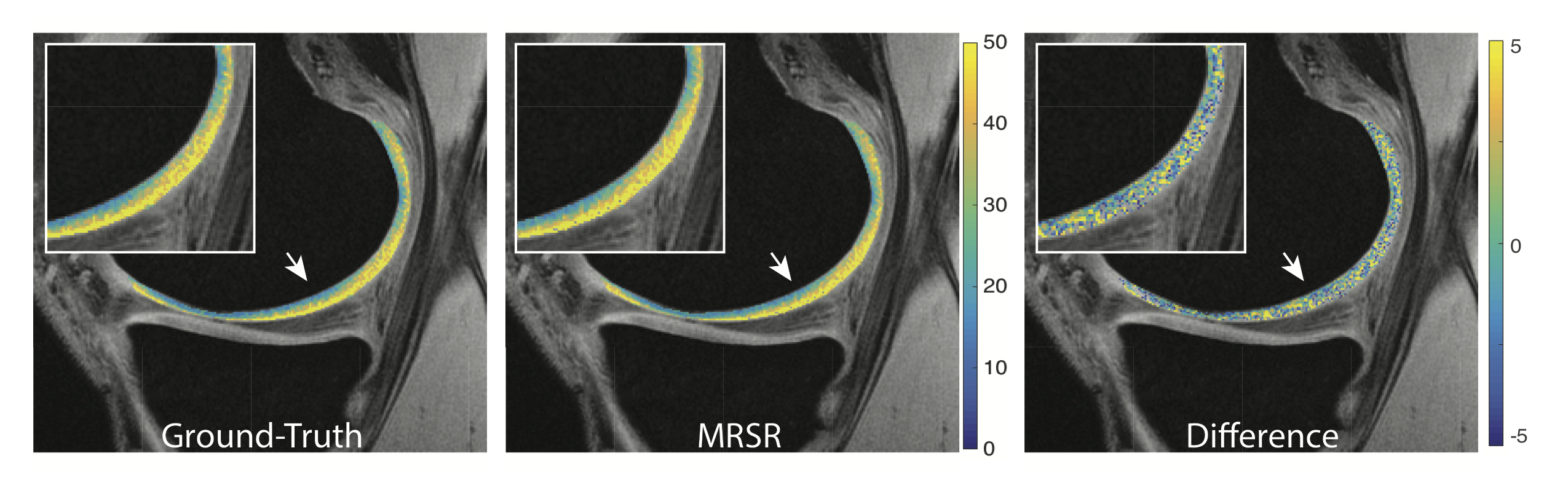} 
\end{center}
\vspace*{-5mm}
\caption[Example MRSR T$_2$ relaxation time maps compared to the ground-truth]{MRSR T$_2$ relaxation time maps appear similar and provide a similar spatial distribution of T$_2$ values compared to the ground-truth. The difference map has no discernible structure, suggesting minimal systematic bias. (note the different color scale).}
\label{fig:t2}
\end{figure}

Example coronal and axial images of the resolution enhancement are shown in Fig.\ref{fig:cor} and Fig.\ref{fig:ax}. The medial collateral ligament (solid arrow, approximately 1mm thick) is completely blurred out in the input image (Fig. \ref{fig:cor}), but can be delineated well with MRSR. Similarly, the ligament bundles (dashed arrow) and the synovium (dotted arrow) appeared blurrier in the input image than the MRSR. Fig. \ref{fig:ax} shows that signal irregularities in medial synovium (solid arrow) delineated better using MRSR than in the input image. The lateral synovial membrane (dotted arrow) also appears thickened in the blurred input image but not in the ground-truth or MRSR, which may incorrectly lead to a diagnosis of synovitis. The patellar cartilage (dashed arrow) appears blurred with diffuse signal heterogeneity in the input image, which may lead to an incorrect cartilage lesion diagnosis. Example T$_2$ map comparisons (shown in Fig.\ref{fig:t2}) show minimal differences between the ground-truth and MRSR images, and that the per-pixel difference map has no organized structure, suggesting minimal systematic bias.

\begin{table}[t]
\centering
\caption{Quantitative image quality metrics for both DESS echoes comparing the ground-truth to MRSR, TCI, FI, and ScSR images for test cohort 1. * indicates a significant difference (p$<$0.05) compared to MRSR. $^{\dagger}$ indicates that all displayed values are multiplied by 10$^{3}$.}
\vspace*{-4mm}
\label{tab:morph}
\begin{center}
\renewcommand{\arraystretch}{1.25}

\begin{tabular}[h]
{>
{\centering\arraybackslash}m{0.60in} | >{\centering\arraybackslash}p{0.55in} | >{\centering\arraybackslash}p{0.85in} | >{\centering\arraybackslash}p{0.85in} | >{\centering\arraybackslash}p{0.85in} | >{\centering\arraybackslash}p{0.85in}}
\hline
\textbf{Metric} & \textbf{Image} & \textbf{MRSR} & \textbf{TCI} & \textbf{FI} & \textbf{ScSR} \\
\hline

\multirow{2}{*}{SSIM} &  S$_1$  &  0.98 $\pm$ 0.01  &  0.95 $\pm$ 0.02*   &  0.92 $\pm$ 0.02*   &  0.97 $\pm$ 0.01* \\
					  &  S$_2$  &  0.98 $\pm$ 0.01  &  0.96 $\pm$ 0.02*   &  0.95 $\pm$ 0.02*   &  0.97 $\pm$ 0.01 \\
\hline           
           
\multirow{2}{*}{pSNR} &  S$_1$  &  37.7 $\pm$ 1.5   &  32.5 $\pm$ 3.6*    &  32.4 $\pm$ 2.8*    &  36.6 $\pm$ 1.1  \\
					  &  S$_2$  &  38.7 $\pm$ 2.0   &  33.6 $\pm$ 4.2     &  33.6 $\pm$ 3.5*    &  37.5 $\pm$ 1.6  \\
\hline           
           
\multirow{2}{*}{RMSE$^{\dagger}$} &  S$_1$  &  0.18 $\pm$ 0.06  &  0.72 $\pm$ 0.56*   &  0.69 $\pm$ 0.47*   &  0.22 $\pm$ 0.05 \\
					  &  S$_2$  &  0.13 $\pm$ 0.04  &  0.51 $\pm$ 0.40    &  0.47 $\pm$ 0.34*   &  0.16 $\pm$ 0.05 \\
\hline  

\end{tabular}
\end{center}
\end{table}
\vspace*{-10mm}

\begin{table}[t]
\centering
\caption{Cartilage T$_2$ relaxation times for MRSR, TCI, FI, and ScSR compared to the ground-truth using differences and coefficients of variation (CV\%) in test cohort 2.}
\vspace*{-4mm}
\label{tab:t2}
\begin{center}
\renewcommand{\arraystretch}{1.25}
\begin{tabular}[h]{  >
{\centering\arraybackslash}p{0.7in} | >
{\centering\arraybackslash}p{0.78in} | >
{\centering\arraybackslash}p{0.78in} | >
{\centering\arraybackslash}p{0.78in} | >
{\centering\arraybackslash}p{0.78in} | >
{\centering\arraybackslash}p{0.78in}
}
\hline
\textbf{Subject} & \textbf{Ground-Truth} & \textbf{MRSR} & \textbf{TCI}  & \textbf{FI}  & \textbf{ScSR}\\
\hline

1  &  35.2  &  35.8 &  36.4  &  36.1 &  42.4  \\
2  &  42.6  &  44.1 &  44.4  &  44.5 &  50.1  \\
3  &  27.9  &  29.1 &  29.8  &  29.4 &  35.9  \\
4  &  35.3  &  38.5 &  39.5  &  39.0 &  58.3  \\
5  &  36.6  &  38.0 &  39.0  &  39.2 &  46.7  \\

\hline\rule{0pt}{13pt}
\textbf{Average}   & \textbf{35.5$\pm$5.2} & \textbf{37.1$\pm$5.4} & \textbf{37.8$\pm$5.3} & \textbf{37.6$\pm$5.5} & \textbf{46.7$\pm$8.4} \\ 

\textbf{CV \%}   & N/A  & \textbf{3.1$\pm$1.8} & \textbf{4.5$\pm$2.2} & \textbf{4.1$\pm$2.0} & \textbf{18.8$\pm$9.3}\\ 

\textbf{Difference}  & N/A & \textbf{1.6$\pm$1.0} & \textbf{2.3$\pm$1.1} & \textbf{2.1$\pm$1.1} & \textbf{11.2$\pm$6.7}\\ 

\textbf{CCC}  & N/A & \textbf{0.93} & \textbf{0.87} & \textbf{0.89} & \textbf{0.21}\\ 

\end{tabular}
\end{center}
\end{table}
\vspace*{6mm}

%
\section{Discussion and Conclusion}

In this study, we demonstrated that transfer learning can be effectively used to perform SR on MRI sequences with varied contrasts that are used clinically and in epidemiological studies, even with a small training dataset. The dual-contrast DESS sequence was able to maintain a considerably higher resolution and detail than the comparison methods. It is important to note that since the SR was carried out only in one dimension of the 3D dataset, the image enhancements in Fig.\ref{fig:cor} and Fig.\ref{fig:ax} are more prominent in the left-right direction anatomically, which is also the same direction of the displayed images. 

The MRSR approach maintained comparable T$_2$ relaxation times between the ground-truth. A pixel-wise CV of 3\% has shown to be adequate for use in OA studies and a CCC of over 0.90 indicated excellent reproducibility compared to the ground-truth \cite{Baum2013}. With MRSR, slices can be acquired with a higher section thickness for accurate T$_2$ measurement, while enabling super-resolution for performing high-resolution MRI scans, which was not possible previously due to SNR limitations. Interestingly enough, all methods over-estimated T$_2$ values, likely because the thin cartilage has two major divisions (deep and superficial), where the deep cartilage has lower signal. Blurring from the superficial cartilage would increase signal in the deeper layer, leading to a higher T$_2$ value. Performing layer-wise T$_2$ values will be important in future studies. 

In conclusion, we demonstrated how SR enhanced through-plane resolution in MRI and maintained quantitative accuracy of the T$_2$ relaxation time biomarker. MRSR outperforms conventional and state-of-the-art resolution enhancement methods and has potential for use in clinical and research studies.

\bibliography{refs}
\bibliographystyle{splncs}

\end{document}